\documentclass[graybox]{svmult}

\usepackage{helvet}         % selects Helvetica as sans-serif font
\usepackage{courier}        % selects Courier as typewriter font
\usepackage{type1cm}        % activate if the above 3 fonts are
\usepackage{makeidx}         % allows index generation
\usepackage{graphicx}        % standard LaTeX graphics tool
\usepackage{multicol}        % used for the two-column index
\usepackage[bottom]{footmisc}% places footnotes at page bottom

\usepackage{cite}
\usepackage{amsmath,amssymb,amsfonts}
\usepackage{algorithmic}
\usepackage{graphicx}
\usepackage{textcomp}
\usepackage{xcolor}

\usepackage{microtype}
\usepackage{paralist}
\usepackage{multirow}
\usepackage{graphicx}
\usepackage{bm}
\usepackage{cite}
\usepackage{amsmath}
\usepackage{color}
\newcommand{\red}{\color{red}}
\usepackage[switch]{lineno}
\usepackage{pgfplotstable}
\usepackage{pgfplots}
\pgfplotsset{compat=1.16}
\usepackage{subcaption}
\usepackage{adjustbox}
\RequirePackage{filecontents} 

\makeindex             % used for the subject index
\begin{document}

% \title*{Contribution Title}
\title*{Out-of-Scope Domain and Intent Classification through Hierarchical Joint Modeling}
\titlerunning{Out-of-Scope Domain and Intent Classification}
% \titlerunning*{Contribution Title}
\author{Pengfei Liu\thanks{This paper was published in the 12th International Workshop on Spoken Dialog System Technology (IWSDS), 2021. Pengfei Liu is the corresponding author.}, Kun Li and Helen Meng}

\institute{Pengfei Liu \at SpeechX Limited, Shenzhen, China, \email{pfliu@speechx.cn}
\and Kun Li \at SpeechX Limited, Shenzhen, China, \email{kli@speechx.cn}
\and Helen Meng \at The Chinese University of Hong Kong, Hong Kong, China, \\ \email{hmmeng@se.cuhk.edu.hk}
}

\maketitle

\abstract{
User queries for a real-world dialog system may sometimes fall outside the scope of the system's capabilities, but appropriate system responses will enable smooth processing throughout the human-computer interaction. This paper is concerned with the user's intent, and focuses on out-of-scope intent classification in dialog systems. Although user intents are highly correlated with the application domain, few studies have exploited such correlations for intent classification. Rather than developing a two-stage approach that first classifies the domain and then the intent, we propose a hierarchical multi-task learning approach based on a joint model to classify domain and intent simultaneously. Novelties in the proposed approach include: (1) sharing supervised out-of-scope signals in joint modeling of domain and intent classification to replace a two-stage pipeline; and (2) introducing a hierarchical model that learns the intent and domain representations in the higher and lower layers respectively. Experiments show that the model outperforms existing methods in terms of accuracy, out-of-scope recall and $F_1$. Additionally, threshold-based post-processing further improves performance by balancing precision and recall in intent classification.
}

%%%%%%%%%%%%%%%%%%%%
%%%%%%%%%%%%%%%%%%%%%%
\section{Introduction}

%%%%%%%%%%%%%%%%%%%%%%
% What is the problem?
%%%%%%%%%%%%%%%%%%%%%%
%% Background
Intent classification \cite{liu2019review} is one of the core components for NLU in dialog systems, where NLU needs to recognize the domain, intent and slots of a user query to make an appropriate response.
Out-of-scope user queries are inevitable in a task-oriented dialog system, because it is difficult, if not impossible, to convey precisely and comprehensively to the system the range of capabilities of the system, especially in terms of the supported intents \cite{larson2019evaluation}.
However, the problem of out-of-scope intent classification, which aims to find out the queries not belonging to any of the system-supported intents, is not so actively investigated due to lack of publicly available datasets.
%% Problem statement
% This problem aims to find out the queries which users may reasonably make, but fall outside the scope of the system-supported intents \cite{larson2019evaluation}. 
This problem is similar to out-of-distribution intent classification \cite{hendrycks2017baseline, lin2019deep}, but poses new challenges since the out-of-scope queries are often similar with the in-scope queries, in terms of topics and/or styles \cite{larson2019evaluation}.
Existing approaches to out-of-scope intent classification include: (1) two-step approaches which first perform binary classification of in-scope versus out-of-scope and in the former case further classify the specific in-scope intent \cite{larson2019evaluation,lin2019deep}; (2) classifier-based approaches that place out-of-scope query as an additional intent category \cite{larson2019evaluation, wu2020tod}; and further extend this with (3) a threshold for classification probabilities for each in-scope intent and optionally augmented with an out-of-scope intent \cite{meng1999believe,larson2019evaluation,lin2019post}.

%%%%%%%%%%%%%%%%%%%%%%%%%%%%%%%%
% Why? What are the motivations?
%%%%%%%%%%%%%%%%%%%%%%%%%%%%%%%%
As can be seen, out-of-scope intent classification has not yet been studied from the perspective of joint modeling or multi-task learning.
Intent classification in dialog systems are highly dependent on supported domains, such as banking, restaurant, shopping etc., which means that domain information is useful for recognizing the intent of a user query.
Although there has been studies on joint models for the tasks of intent classification and slot filling \cite{xu2013convolutional,guo2014joint,liu2016attention,zhang2016joint,kim2017onenet,goo2018slot,wang2018bi,chen2019bert,zhang2019joint}, multi-task joint modeling of domains and intents are rarely studied \cite{hakkani2016multi,kim2017onenet,kim2018joint}.
% Furthermore, there are still lack of studies to explicitly model the hierarchical relationships among different NLP tasks in the multi-task learning framework.
Furthermore, there still lacks deep understanding of the settings in which multi-task learning may bring significant benefits \cite{sanh2019hierarchical}, in other words, how to effectively model the correlation between domain and intent classification in a multi-task learning framework. 
% Although multi-task learning is effective in a lot of applications, there is still lack of studies to explicitly model the relationships among different tasks.
Remarkably, \cite{sanh2019hierarchical} introduced a hierarchical multi-task learning model for a set of carefully selected semantic tasks, aiming to supervise lower-level tasks (e.g., NER) at the bottom layers and more complex tasks (e.g., relation extraction) at the top layers of the model. %including Named Entity Recognition, Entity Mention Detection, Coreference Resolution and Relation Extraction,

%%%%%%%%%%%%%%%%%%%%%%%%%%%%%%%%%%%
% How do you deal with the problem?
%%%%%%%%%%%%%%%%%%%%%%%%%%%%%%%%%%%
This paper presents a hierarchical joint model for out-of-scope domain and intent classification, where the two tasks of domain and intent classification share the same out-of-scope supervised signals through joint modeling, and a hierarchical structure is introduced in the network to learn the intent representation on top of the domain representation.
% the joint model and a hierarchical structure is introduced in the network to learn the intent representation on top of the domain representation.
The major benefits of joint modeling and hierarchical structure are \textit{information sharing and inheritance} between domain and intent classification, which may present advantages over a two-stage pipeline approach of domain classification followed by intent classification.
% Joint modeling enables to share the supervised signals of a user query between domain and intent classification. 
The motivation to introduce the hierarchical structure in the network are two-fold: (1) there are generally a larger number of intents than domains and consequently intent classification may need a more refined semantic understanding of the user's query than domain classification; and (2) intent classification can generally benefit from additional domain-related information.
% (1) intent classification needs a model with more layers than domain classification since intents have a larger number of classes than domains and may require deeper understanding of the semantics of a user query; (2) additional domain information is helpful for intent classification.

For example, in the user query of ``\textit{My credit card was swallowed by ATM when I tried to withdraw some money. How can I get back my card?}" --- it is easy to determine the domain as \textit{banking} based on the words like \textit{credit card} or \textit{ATM}, but requires a model of more refined understanding to determine that the intent is ``\textit{report card swallowed}" instead of ``\textit{withdraw money}". 
Besides, knowing that the domain of the query is in banking gives additional information for intent classification.
From the perspective of representation learning, the proposed joint model introduces a hierarchical bias whereby the higher layers represent intent information, while the lower layers represent domain information. 
Such an organization offer a better knowledge representation than a flat structure shared between domain and intent.  The major contributions of this paper are:
%%%%%%%%%%%%%%%%%%%%%%%%%%%%%%
% What are your contributions?
%%%%%%%%%%%%%%%%%%%%%%%%%%%%%%
%% Summarize contributions
% We summarize the major contributions in this paper as follows:
\begin{enumerate}[(1)]\setlength{\itemsep}{0pt}
    \item We propose a novel multi-task joint model for out-of-scope domain and intent classification, which outperforms state-of-the-art methods by a large margin; % due to sharing the out-of-scope supervised signals between the tasks;
    \item We introduce a hierarchical structure in the model to allow for hierarchical representation learning and information inheritance from domain to intent;
    % \item We show that a simple threshold-based method on the \texttt{softmax} outputs of the network improves the performance further and provides an effective way to balance precision and recall in out-of-scope intent classification.
    \item We show that a threshold-based post-processing method improves the performance further by balancing precision and recall in out-of-scope intent classification.
    % \item We demonstrate that the proposed model is also effective in the setting of few-shot learning.
\end{enumerate}

%%%%%%%%%%%%%%%%%%%%

%%%%%%%%%%%%%%%
%%%%%%%%%%%%%%%%%%%%%%
\section{Related Work}

%% out-of-scope intent classification problem
The problem of out-of-scope intent classification is not as actively studied due to lack of publicly available datasets \cite{braun2017evaluating,coucke2018snips,liu2019benchmarking,larson2019evaluation,yilmaz2020kloos}, but is nonetheless very important especially in real-world dialog systems \cite{aliannejadi2019asking}.
The out-of-scope problem encompasses cases where the intent of a user query is not supported by the dialog system but the query is similar in style and or topic to the in-scope queries, as is reflected by the term \textit{out-of-scope} \cite{larson2019evaluation,wu2020tod}. It also encompasses cases where the user query originates from another dataset and is substantially different from the in-distribution queries, which is literally an \textit{out-of-distribution} problem \cite{hendrycks2017baseline,lin2019deep}.
% out-of-scope classification
For out-of-scope intent classification, \cite{larson2019evaluation} introduced a 150-intent dataset for evaluating out-of-scope prediction performance of intent classification systems, and presented BERT-based methods which are however poor at recognizing out-of-scope intents.
In line with this formulation, \cite{wu2020tod} introduced a pre-trained language model named ToD-BERT which is learned from a bunch of task-oriented dialogue datasets and obtained better performance than BERT in terms of accuracy and out-of-scope recall on a downstream intent classification task.
% out-of-distribution detection
By contrast, \cite{hendrycks2017baseline} studied the out-of-distribution problem by forming out-of-distribution examples from another dataset and found that classification with \textit{softmax} distribution probabilities offer good performance on out-of-distribution detection.
Similarly, \cite{lin2019deep} considered the intents excluded from the training set as out-of-distribution intents and adopted a novelty detection algorithm named \textit{local outlier factor} to detect the unknown intents.

%% Differences
This paper presents a novel approach for out-of-scope intent classification based on joint modeling of domain and intent, together with hierarchical representation fine-tuning from the BERT model for the correlated tasks of domain and intent classification.
Hierarchical multi-task learning has been introduced for semantic tasks such as named entity recognition, entity mention detection, coreference resolution and relation extraction \cite{sanh2019hierarchical}. Similarly, hierarchical modeling has also been applied to syntactic and semantic tasks in chunking, dependency parsing, semantic relatedness, and textual entailment \cite{hashimoto2017joint}. To the best of our knowledge, the present work is the first to apply hierarchical joint modeling to out-of-scope domain and intent classification.

%%%%%%%%%%%%%%%

%%%%%%%%%%%%%%
%%%%%%%%%%%%%%%%%%%%%
% \section{Methodology} \label{sec:method}
\section{Hierarchical Joint Modeling} \label{sec:method}
% This section first explains the pre-trained BERT model \cite{devlin2019bert} for utterance (i.e., user query) representation, and then introduces the proposed hierarchical joint model, named BERT-Joint, for joint domain and intent classification in a multi-task learning framework.

\begin{figure}[htb]
  \centering
   \includegraphics[width=0.61\linewidth]{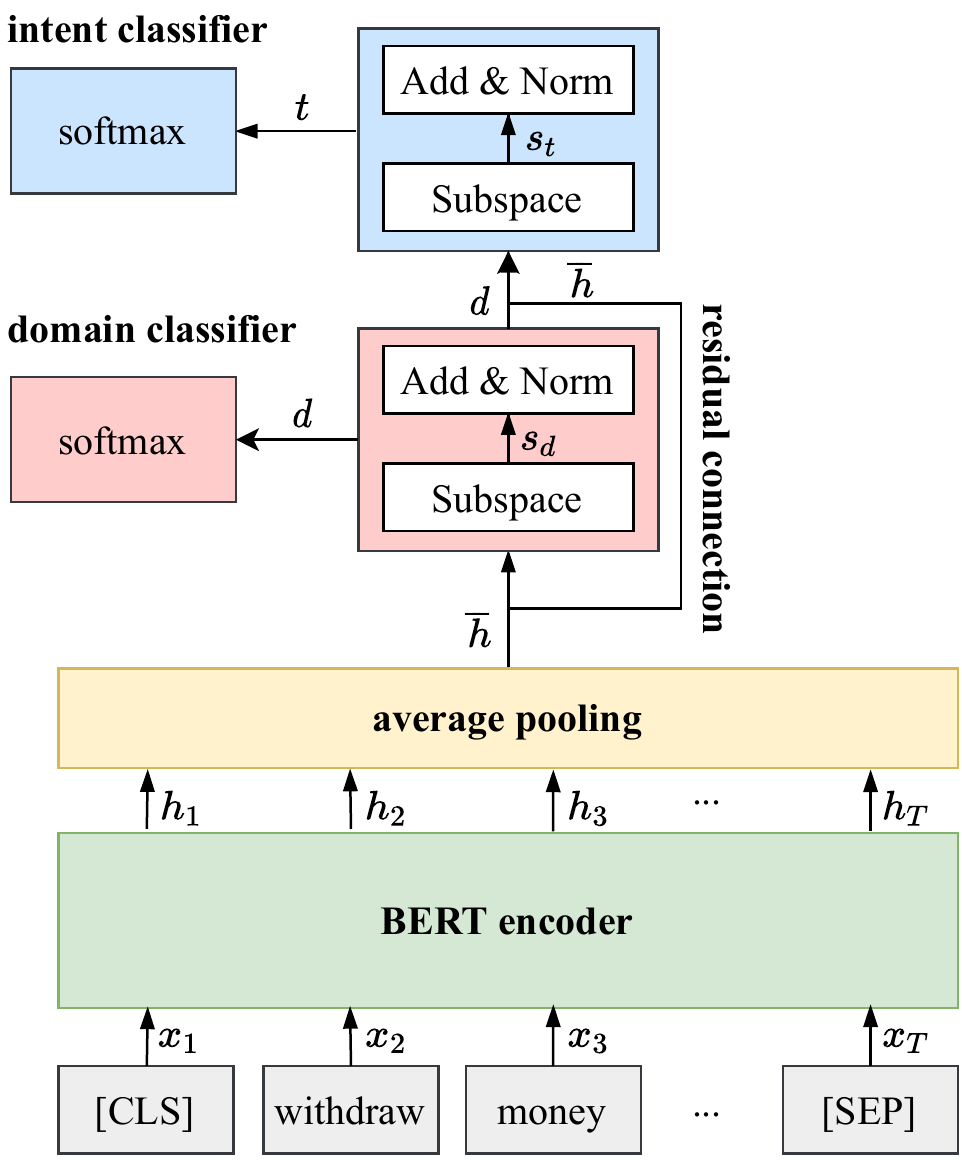}
%   \caption{The architecture of the proposed hierarchical BERT-based joint model, where Trm is the transformer layer \cite{vaswani2017attention}.}
%   \vspace{-0.5em}
  \caption{The architecture of BERT-Joint model, where a pre-trained BERT model is adopted as the encoder, \texttt{[CLS]} and \texttt{[SEP]} are the two special tokens adding to the start and the end of each sequence respectively.}
  \label{fig:model}
\end{figure}

% We propose a hierarchical joint model, named BERT-Joint, for joint domain and intent classification in a multi-task learning framework.
The proposed hierarchical joint model, named BERT-Joint, is illustrated in Figure~\ref{fig:model}, where a token sequence is fed into a BERT encoder to obtain a sequence of hidden states that are averaged by a pooling operation to obtain the BERT representation $\bar{h}$. The following modules are a domain encoder in red and an intent encoder in blue, as well as the subsequent \texttt{softmax} layers for domain and intent classification respectively.
Particularly, the intent encoder is fed with the domain representation $d$ to model the hypothesis that intent classification needs additional domain information and requires more layers than domain classification to learn the intent representation $t$.
% Note that the intent encoder is also fed with the domain representation to model our hypothesis that knowing the domain of an utterance is helpful for intent classification.
% \vspace{-0.5em}

%%%%%%%%%%%%%%%%%%%%%%%%%%%%%%%%%%%%%%%%
\noindent{\textbf{BERT Representation.}}
For a given utterance, BERT \cite{devlin2019bert} takes the word sequence $x=(x_1, x_2, ..., x_T)$ as input, and outputs a sequence of hidden states $h=(h_1, h_2, ..., h_T)$ after a few Transformer layers.
Following the training schema in pre-trained BERT models, a special token \texttt{[CLS]} is added to the start of every sequence for aggregating the sequence representation and another special token \texttt{[SEP]} is appended to the end of every sequence for differentiating the sentences \cite{devlin2019bert} . 
We used the average pooling vector $\bar{h} = \texttt{average-pooling}(h)$, as the BERT representation of an utterance, which gives slightly better performance than $h_1$ corresponding to \texttt{[CLS]}.

%%%%%%%%%%%%%%%%%%%%%%%%%%%%%%%%%%%%%%%%%%
\noindent{\textbf{Domain Representation.}}
Given an utterance representation $\bar{h}$ from the BERT encoder, we first obtain a representation subspace $s_d$ from $\bar{h}$ using a non-linear transformation with the weight matrix $W_d$ and the additive bias $b_d$, and then apply residual connection \cite{he2016deep} and layer normalization \cite{ba2016layer} to obtain the domain representation vector $d$, as illustrated in Equation (1) and (2), which are inspired from the Transformer model \cite{vaswani2017attention}.
\begin{align}
    s_d &= \texttt{ReLU}(W_d\bar{h} + b_d) \\
    d &= \texttt{LayerNorm}(s_d + \bar{h})
\end{align}
% Since both domain and intent share the same BERT representation at the lowest layer in the joint model, we assume that they have different subspace vectors from the BERT representation. Equation (1) and (2) are inspired from the Transformer model \cite{vaswani2017attention} to get a subspace vector from $\bar{h}$.

%%%%%%%%%%%%%%%%%%%%%%%%%%%%%%%%%%%%%%%%%%
\noindent{\textbf{Intent Representation.}}
Similarly, we obtain a representation subspace $s_t$, as in Equation (3), for intent transformed from the summation of the domain representation $d$ and the BERT representation $\bar{h}$, where $W_t$ is the weight matrix and $b_t$ is the additive bias. We then apply residual connection and layer normalization to get the intent representation $t$ in Equation (4).
\begin{align}
    s_t &= \texttt{ReLU}(W_t(d + \bar{h}) + b_t) \\
    t &= \texttt{LayerNorm}(s_t + d)
\end{align}
The intent representation $t$ is built on top of the domain representation $d$ to introduce a hierarchical structure in the network.
% Such bias aims to model the hypothesis that domain classification relies on the representation learned from the lower-level layers in the network while intent classification needs the more complex representation obtained with more layers.
Such hierarchical structure aims to capture the dependency between a domain and the corresponding intents, and model the hypothesis that additional domain information is useful for intent classification. Besides, we believe that intent classification needs a model with more layers than domain classification due to a larger number of intent classes and the requirement for a deeper understanding of the utterance semantic.
% Other methods like concatenation of $d$ and $\bar{h}$ can also be used to incorporate the domain representation. %, although giving slightly worse performance than summation in our experiments.

%%%%%%%%%%%%%%%%%%%%%%%%%%%%%%%%%%%
\noindent{\textbf{Joint Learning.}}
We learn domain and intent classification jointly using two separate \texttt{softmax} layers on top of the corresponding representations, as illustrated in Equation (5-6), where $p^d$ is the predicted domain distribution and $p^t$ is the predicted intent distribution.
We adopt the cross entropy loss for model training. $L_d$ is the loss between the predicted domain distribution $p^d$ and the true domain $y^d$, where $p_m^d$ means the predicted probability of being domain $m$, and $y_m^d$ is 1 if the true domain is $m$ else 0. Similarly, $L_t$ measures the loss between the predicted intent distributions $p^t$ and the true intent $y^t$, where $p_n^t$ means the predicted probability that of being intent $n$. $y_n^t$ is also a binary indicator which is 1 if the true intent is $n$ and else 0.
We optimize domain and intent classification jointly using a linear combination of their corresponding cross entropy loss, as shown in Equation (9), where the weight $\lambda$ is also a learnable parameter, jointly learned with the other model parameters. %, e.g., $W^d, b^d, W^t, b^t$, etc. %, such as $W^d, b^d, W^t, b^t$, etc.
\begin{align}
    p^d &= \texttt{softmax}(W^d d + b^d) \\
    p^t &= \texttt{softmax}(W^t t + b^t) \\
    % L_d & = -\frac{1}{N} \log y^d \\
    %  L_t & = -\frac{1}{N} \log y^t \\
    L_d &= -\sum_{m=1}^M y_m^d \log(p_m^d) \\
    L_t &= -\sum_{n=1}^N y_n^t \log(p_n^t) \\
    L &= \lambda L_d + (1-\lambda) L_t
\end{align}

Note that the number of domains $M$ is much smaller than the number of intents $N$ in real-world dialog systems, which means that it is easier to determine the domain of an utterance than the intent.
As each utterance has both labels of domain and intent, the advantage of joint learning is that the discrimination capability learned by the domain classifier, particularly on out-of-scope user queries, are also shared with the intent classifier by feeding the domain representation to the subsequent intent representation layers.

%%%%%%%%%%%%%%%%%%%%%%%%%%%%%%%%%%%%%%%%%%%%%%%%%%%%
\noindent{\textbf{Threshold-based Post-processing.}}
Since out-of-scope examples are frequently misclassified as in-scope intents at low probabilities, we propose a threshold-based method to post-process the predicted probabilities, and consider an example as out-of-scope if the predicted probability is below the pre-specified threshold $\tau$, (i.e., $p^t < \tau$ for intent classification).
It is interesting to observe that setting a threshold value generally improves both in-scope and out-of-scope accuracy.
% (See Table~\ref{tab:results}).
More importantly, the threshold-based post-processing method provides an effective way to balance \textit{precision} and \textit{recall} for out-of-scope intent classification.

%%%%%%%%%%%%%%

%%%%%%%%%%%%%%%%%%%
%%%%%%%%%%%%%%%%%%%%%
\section{Experiments}

%%%%%%%%%%%%%%%%%%%%%%%%%%%%%%%
\subsection{Experimental Setup}

%%%%%%%%%%%%%%%%%%%%%%%%%%%%
\noindent{\textbf{Dataset.}}
We evaluate the proposed model using the OOS dataset \cite{larson2019evaluation}, which consists of 150 intents across 10 domains and a number of out-of-scope examples belonging to none of the domains or intents\footnote{https://github.com/clinc/oos-eval}.
The dataset is different from conventional intent datasets in the sense that it focuses on out-of-scope intent classification. The task is particularly challenging since the out-of-scope examples are similar in topics or styles with the in-scope examples but are not within any of the 150 in-scope intents.
There are three variants of the OOS dataset, namely \textit{Small}, \textit{Imbalanced} and \textit{OOS+}, where \textit{Small} has the smallest number of total examples, and \textit{OOS+} has the largest number of out-of-scope examples. In contrast, \textit{Imbalanced} has the imbalanced number of in-scope examples.
% to represent different production scenarios.
The number of examples in each variant of the OOS dataset is shown in Table~\ref{tab:dataset}.
Note that all the variants have the same test set which has 1000 out-of-scope examples and 150 * 30 in-scope examples.
\begin{table}[htb]
\centering
\caption{Number of examples in variants of the OOS dataset.}
\label{tab:dataset}
\resizebox{\linewidth}{!}{%
\begin{tabular}{r|r|r|r|r|r}
\hline
 &  & \textbf{Full} & \textbf{Small} & \textbf{Imbalanced} & \textbf{OOS+} \\ \hline
\multirow{3}{*}{\textbf{Train}} & \textbf{Total Examples} & 15100 & 7600 & 10625 & 15250 \\
 & \textbf{\#Out-of-scope Examples} & 100 & 100 & 100 & 250 \\
 & \textbf{\#Examples per In-scope Intent} & 100 & 50 & 25, 50, 75, 100 & 100 \\ \hline
\multirow{3}{*}{\textbf{Valid}} & \textbf{Total Examples} & 3100 & 3100 & 3100 & 3100 \\
 & \textbf{\#Out-of-scope Examples} & 100 & 100 & 100 & 100 \\
 & \textbf{\#Examples per In-scope Intent} & 20 & 20 & 20 & 20 \\ \hline
\multirow{3}{*}{\textbf{Test}} & \textbf{Total Examples} & 5500 & 5500 & 5500 & 5500 \\
 & \textbf{\#Out-of-scope Examples} & 1000 & 1000 & 1000 & 1000 \\
 & \textbf{\#Examples per In-scope Intent} & 30 & 30 & 30 & 30 \\ \hline
\end{tabular}%
}
\end{table}

%%%%%%%%%%%%%%%%%%%%%%%%%%%%
\noindent{\textbf{Metrics.}}
We adopt \textit{accuracy} as the metric for evaluating the overall accuracy (all) on all the examples, and the in-scope accuracy (in) on the in-scope examples, for the OOS test set. For out-of-scope examples, we report the metrics of precision ($P$), recall ($R$) and $F_1$.

%%%%%%%%%%%%%%%%%%%%%%%%%%%%%%%%%%%%%%%
\noindent{\textbf{Settings.}}
We adopt the pre-trained BERT model of \textit{bert-base-uncased} for an initial utterance representation.
For fine-tuning, we used the AdamW \cite{loshchilov2017decoupled} optimizer and set the proportion of warm-up steps as $0.1$, the learning rate as \texttt{4E-5}. The maximum number of epochs is set as 10 on all the experiments except on OOS+, which has the largest number of training examples and obtains the best performance using 5 epochs. 
% If the intent classification accuracy does not increase for 3 epochs, we stop the fine-tuning procedure earlier.
% To speed up training and avoid overfitting, 
We adopted early stopping on condition that the intent classification accuracy does not improve for 3 epochs.
% We early stopped the fine-tuning procedure if the intent classification accuracy does not increase for 3 epochs.
% We set the dropout rate as $0.1$ for the BERT related layers and all other layers as $0.5$. 
We implemented the models using the PyTorch framework \cite{paszke2019pytorch} and kept the random seed fixed on all the experiments for reproducible results.
% For the threshold-based post-processing on \textit{softmax} outputs, we chose the threshold value having the highest overall accuracy on the validation dataset with a grid search over $\{0.1, 0.2, \dots, 0.9\}$.

%%%%%%%%%%%%%%%%%%%%%%%%%%%%%%%%%%%
\subsection{Results and Discussion}

%%%%%%%%%%%%%%%%%%%%%%%%%%%%%%%%%%%%%%%%%%%%%%%%%%%%%%
\noindent{\textbf{Comparisons with Existing Methods.}}
Table~\ref{tab:results} presents the experimental results from \cite{larson2019evaluation} including the methods of FastText, SVM, CNN and BERT, and \cite{wu2020tod} covering GPT2, DialogGPT and ToD-BERT, as well our methods of BERT and BERT-Joint. It can be seen that the proposed BERT-Joint model obtains the best performance in terms of overall accuracy, and out-of-scope precision ($P$), recall ($R$) and $F_1$, and is further outperformed by applying a threshold-based post-processing method.

\begin{table}[htb]
\centering
\caption{Performance comparisons with existing methods for intent classification on the OOS test dataset (Full).}
% \vspace{-0.5em}
\label{tab:results}
\resizebox{0.92\linewidth}{!}{%
\begin{tabular}{r|l|cc|ccc}
\hline
\multirow{2}{*}{} & \multirow{2}{*}{\textbf{Model}} & \multicolumn{2}{c|}{\textbf{Accuracy}} & \textbf{$P$} & \textbf{$R$} & \textbf{$F_1$} \\ \cline{3-7} 
 &  & \textbf{all} & \textbf{in} & \textbf{out} & \textbf{out} & \textbf{out} \\ \hline
\multirow{5}{*}{\textbf{Larson et al. \cite{larson2019evaluation}}} & \textbf{FastText} & - & 0.890 & - & 0.097 & - \\
 & \textbf{SVM} & - & 0.910 & - & 0.145 & - \\
 & \textbf{CNN} & - & 0.912 & - & 0.189 & - \\ %\cline{2-8} 
 & \textbf{BERT} & - & \textbf{0.969} & - & 0.403 & - \\ \hline
%  & \multicolumn{1}{r|}{+Threshold} & - & 0.962 & - & - & 0.523 & - \\ \hline
\multirow{5}{*}{\textbf{Wu et al. \cite{wu2020tod}}} & \textbf{GPT2} & 0.830 & 0.941 & - & 0.320 & - \\
 & \textbf{DialoGPT} & 0.839 & 0.955 & - & 0.321 & - \\
 & \textbf{BERT} & 0.849 & 0.958 & - & 0.356 & - \\
 & \textbf{ToD-BERT-mlm} & 0.859 & 0.961 & - & 0.463 & - \\
 & \textbf{ToD-BERT-jnt} & 0.866 & 0.962 & - & 0.436 & - \\ \hline
 \multirow{3}{*}{\textbf{This Work}} & \textbf{BERT} & 0.855 & 0.962 & 0.981 & 0.370 & 0.537 \\
 & \textbf{BERT-Joint} & \textbf{0.876} & 0.964 & \textbf{0.984} & \textbf{0.484} & \textbf{0.649} \\ \cline{2-7}
 & \multicolumn{1}{r|}{\textbf{+Threshold}} & \textbf{0.920} & 0.955 & 0.902 & \textbf{0.761} & \textbf{0.825} \\ \hline
\end{tabular}%
}
\end{table}

%%%%%%%%%%%%%%%%%%%%%%%%%
% Begin of Error Analysis
%%%%%%%%%%%%%%%%%%%%%%%%%
%%%%%%%%%%%%%%%%%%%%%%%%%%%%%%%%%%%
\noindent{\textbf{Error Analysis.}}
We analyzed a few examples from the OOS test set, which are misclassified by either BERT or BERT-Joint, as shown in Table~\ref{tab:case-studies}.
Although Examples 1-4 are out-of-scope, users may naturally ask these questions as they do not precisely know about the system's knowledge scope and capabilities.
BERT-Joint makes correct predictions for Examples 3 and 4 but fails to reject Examples 1 and 2.
Examples 5-7 are quite challenging, as they need a deeper semantic understanding of the sentences such as semantic inference (Example 6), discourse structure (Example 7). BERT-Joint classifies Examples 5 and 6 correctly on both domain and intent but not on Example 7, which actually consists of two sentences and the second sentence delivers the real intent.
We notice that there are some annotation errors on \textit{intent} in Examples 8-10. However, the predicted intents by both BERT and BERT-Joint are reasonable.

\begin{table}[h]
    \caption{Examples from the OOS test set for error analysis, where the labels in red are misclassified. Note that BERT only predicts the intent whereas BERT-Joint predicts both the domain and intent simultaneously.}
    \label{tab:case-studies}
    \begin{subtable}[h]{\linewidth}
        \centering
\caption{Testing Examples.}
\resizebox{0.96\textwidth}{!}{%
\begin{tabular}{r|l|l}
\hline
\multicolumn{1}{c|}{\textbf{ID}} & \multicolumn{1}{c|}{\textbf{Example}} & \multicolumn{1}{c}{\textbf{Domain}} \\ \hline
1 & give me the weather forecast for today & oos \\
2 & how much data does my phone have left this month & oos \\
3 & how many homeless people are there & oos \\
4 & how do i learn more about linguistics & oos \\
5 & i would like to know my vacation days balance & work \\
6 & does bank of america give credit cards to people like me & credit\_cards \\
7 & i'm trying to raise my credit score can you tell me what it is now & credit\_cards \\
8 & someone used my chase card without my authorization & credit\_cards \\
9 & can you call the help desk line for my credit card company & credit\_cards \\
10 & how can i request a new credit card & credit\_cards \\ \hline
\end{tabular}}
\label{tab:example}
\end{subtable}
\hfill
\begin{subtable}[h]{\linewidth}
\centering
\vspace{1em}
\caption{Error Analysis.}
\resizebox{\textwidth}{!}{%
\begin{tabular}{r|l|l|ll}
\hline
\multirow{2}{*}{\textbf{ID}} & \multicolumn{1}{c|}{\textbf{Ground Truth}} & \multicolumn{1}{c}{\textbf{BERT}} & \multicolumn{2}{c}{\textbf{BERT-Joint}} \\ \cline{2-5} 
 & \multicolumn{1}{c|}{\textbf{Intent}} & \multicolumn{1}{c}{\textbf{Intent}} & \multicolumn{1}{c|}{\textbf{Domain}} & \multicolumn{1}{c}{\textbf{Intent}} \\ \hline
1 & oos & \red{weather} & \multicolumn{1}{l|}{utility} & \red{weather} \\
2 & oos & \red{balance} & \multicolumn{1}{l|}{utility} & \red{find\_phone} \\
3 & oos & \red{traffic} & \multicolumn{1}{l|}{oos} & oos \\
4 & oos & \red{translate} & \multicolumn{1}{l|}{oos} & oos \\ \hline
5 & pto\_balance & \red{balance} & \multicolumn{1}{l|}{work} & pto\_balance \\
6 & new\_card & \red{international\_fees} & \multicolumn{1}{l|}{credit\_cards} & new\_card \\
7 & credit\_score & \red{improve\_credit\_score} & \multicolumn{1}{l|}{credit\_cards} & \red{improve\_credit\_score} \\ \hline
8 & report\_lost\_card & \red{report\_fraud} & \multicolumn{1}{l|}{banking} & \red{report\_fraud} \\
9 & replacement\_card\_duration & \red{make\_call} & \multicolumn{1}{l|}{credit\_cards} & \red{make\_call} \\
10 & replacement\_card\_duration & \red{new\_card} & \multicolumn{1}{l|}{credit\_cards} & \red{new\_card} \\ \hline
\end{tabular}}
\label{tab:error}
\end{subtable}
\end{table}

%%%%%%%%%%%%%%%%%%%%%%%%%
% End of Error Analysis
%%%%%%%%%%%%%%%%%%%%%%%%%

\begin{table}[hb]
\centering
\caption{Performance comparisons between BERT and BERT-Joint using different dataset variants of OOS.}
% \vspace{-0.5em}
\label{tab:variants}
\resizebox{0.8\linewidth}{!}{%
\begin{tabular}{l|l|cc|ccc}
\hline
\multirow{2}{*}{} & \multirow{2}{*}{\textbf{Model}} & \multicolumn{2}{c|}{\textbf{Accuracy}} & \textbf{\textbf{$P$}} & \textbf{\textbf{$R$}} & \textbf{\textbf{$F_1$}} \\ \cline{3-7} 
 &  & \textbf{all} & \textbf{in} & \textbf{out} & \textbf{out} & \textbf{out} \\ \hline
\multirow{2}{*}{\textbf{Full}} & \textbf{BERT} & 0.855 & 0.962 & 0.981 & 0.370 & 0.537 \\
 & \textbf{BERT-Joint} & \textbf{0.876} & 0.964 & \textbf{0.984} & \textbf{0.484} & \textbf{0.649} \\ \hline
\multirow{2}{*}{\textbf{Small}} & \textbf{BERT} & 0.845 & 0.953 & 0.975 & 0.357 & 0.523 \\
 & \textbf{BERT-Joint} & \textbf{0.865} & 0.954 & \textbf{0.981} & \textbf{0.464} & \textbf{0.630} \\ \hline
\multirow{2}{*}{\textbf{Imbalanced}} & \textbf{BERT} & 0.855 & 0.952 & \textbf{0.981} & 0.423 & 0.591 \\
 & \textbf{BERT-Joint} & \textbf{0.869} & 0.960 & 0.979 & \textbf{0.462} & \textbf{0.628} \\ \hline
\multirow{2}{*}{\textbf{OOS+}} & \textbf{BERT} & 0.882 & 0.959 & \textbf{0.983} & 0.536 & 0.694 \\
 & \textbf{BERT-Joint} & \textbf{0.897} & 0.959 & 0.969 & \textbf{0.621} & \textbf{0.757} \\ \hline
\end{tabular}%
}
\end{table}

%%%%%%%%%%%%%%%%%%%%%%%%%%%%%%%%%%%%%%%%%%%%%%%%%%%%
\noindent{\textbf{Performance on Dataset Variants.}}
We further verified the performance of BERT-Joint on the OOS variants, as shown in Table~\ref{tab:variants}.
We observe that BERT-Joint consistently outperforms BERT on all the dataset variants in terms of overall accuracy, out-of-scope recall and $F_1$.
Particularly, it improves the $F_1$ score of out-of-scope examples by an absolute increase of more than \textbf{10\%} on \textit{Full} and \textit{Small}, \textbf{3\%} on \textit{Imbalanced} and \textbf{6\%} on \textit{OOS+}.
% Particularly, it improves accuracy, out-of-scope recall and $F_1$ on all on all settings, including \textit{small} and \textit{imbalanced} variants, and increases the $F_1$ score of out-of-scope examples by a large margin.

%%%%%%%%%%%%%%%%%%%%%%%%%%%%%%%%%%%%%

% \noindent{\textbf{Ablation Studies.}}
% \noindent{\textbf{Subspace Layers and Hierarchical Structure.}}
\noindent{\textbf{Effect of Hierarchical Structure.}}
The proposed approach of joint modeling of domain and intent is flexible to support various \textit{flat} or \textit{hierarchical} model structures. Here, \textit{flat} means the domain representation and the intent representation are put side by side in the network, such as $F(\bar{h}; \bar{h})$ which directly uses the same BERT output $\bar{h}$ for domain and intent classification respectively, and $F(s_d; s_t)$ which adopts the subspace vectors $s_d$ and $s_t$ for the corresponding domain and intent classification.
For \textit{hierarchical} model structures, we consider both $H(s_t \rightarrow s_d)$ and $H(s_d \rightarrow s_t)$. The former structure means that we get the intent representation $s_t$ first and then feed it to the subsequent layers to get the domain representation $s_d$, while the latter first learns the domain representation $s_d$ which is then fed to the subsequent layers to get the intent representation $s_t$.

% \vspace{-1em}
\begin{table}[htb]
\centering
% \caption{Performance comparisons between BERT and BERT-Joint with different structures where $F$ means \textit{Flat} and $H$ is short for \textit{Hierarchical}.}
\caption{Performance comparisons between BERT and BERT-Joint with different structures ($F$: \textit{Flat}, $H$: \textit{Hierarchical}).}
% \vspace{-0.5em}
\label{tab:hierarchy}
\resizebox{0.99\linewidth}{!}{%
\begin{tabular}{r|l|l|cc|ccc}
\hline
\multirow{2}{*}{} & \multirow{2}{*}{\textbf{Model}} & \multirow{2}{*}{\textbf{Structure}} & \multicolumn{2}{c|}{\textbf{Accuracy}} & \textbf{$P$} & \textbf{$R$} & \textbf{$F_1$} \\ \cline{4-8} 
 &  &  & \textbf{all} & \textbf{in} & \textbf{out} & \textbf{out} & \textbf{out} \\ \hline
\multirow{5}{*}{\textbf{Full}} & \textbf{BERT} & - & 0.8545 & 0.9622 & 0.9814 & 0.3700 & 0.5374 \\ \cline{2-8} 
 & \multirow{4}{*}{\textbf{BERT-Joint}} & $F(\bar{h}; \bar{h})$ & 0.8689 & 0.9622 & 0.9825 & 0.4490 & 0.6163 \\
 &  & $F(s_d;s_t)$ & 0.8727 & 0.9604 & \textbf{0.9856} & 0.4780 & 0.6438 \\
 &  & $H(s_t \rightarrow s_d)$ & 0.8715 & 0.9611 & 0.9770 & 0.4680 & 0.6329 \\
 &  & $H(s_d \rightarrow s_t)$ & \textbf{0.8764} & \textbf{0.9636} & 0.9837 & \textbf{0.4840} & \textbf{0.6488} \\ \hline
\multirow{5}{*}{\textbf{Small}} & \textbf{BERT} & - & 0.8447 & 0.9531 & 0.9754 & 0.3570 & 0.5227 \\ \cline{2-8} 
 & \multirow{4}{*}{\textbf{BERT-Joint}} & $F(\bar{h}; \bar{h})$ & 0.8529 & 0.9460 & 0.9731 & 0.4340 & 0.6003 \\
 &  & $F(s_d;s_t)$ & 0.8538 & 0.9500 & 0.9768 & 0.4210 & 0.5884 \\
 &  & $H(s_t \rightarrow s_d)$ & 0.8651 & \textbf{0.9573} & \textbf{0.9890} & 0.4500 & 0.6186 \\
 &  & $H(s_d \rightarrow s_t)$ & \textbf{0.8653} & 0.9544 & 0.9810 & \textbf{0.4640} & \textbf{0.6300} \\ \hline
\multirow{5}{*}{\textbf{Imbalanced}} & \textbf{BERT} & - & 0.8555 & 0.9516 & 0.9814 & 0.4230 & 0.5912 \\ \cline{2-8} 
 & \multirow{4}{*}{\textbf{BERT-Joint}} & $F(\bar{h}; \bar{h})$ & 0.8569 & 0.9544 & \textbf{0.9882} & 0.4180 & 0.5875 \\
 &  & $F(s_d;s_t)$ & 0.8673 & 0.9536 & 0.9796 & \textbf{0.4790} & \textbf{0.6434} \\
 &  & $H(s_t \rightarrow s_d)$ & 0.8689 & 0.9587 & 0.9873 & 0.4650 & 0.6322 \\
 &  & $H(s_d \rightarrow s_t)$ & \textbf{0.8693} & \textbf{0.9598} & 0.9788 & 0.4620 & 0.6277 \\ \hline
\multirow{5}{*}{\textbf{OOS+}} & \textbf{BERT} & - & 0.8820 & 0.9589 & 0.9835 & 0.5360 & 0.6939 \\ \cline{2-8}
 & \multirow{4}{*}{\textbf{BERT-Joint}} & $F(\bar{h}; \bar{h})$ & 0.9005 & 0.9600 & 0.9649 & 0.6330 & 0.7645 \\
 &  & $F(s_d;s_t)$ & \textbf{0.9053} & 0.9609 & \textbf{0.9762} & \textbf{0.6550} & \textbf{0.7840} \\
 &  & $H(s_t \rightarrow s_d)$ & 0.8973 & 0.9611 & 0.9744 & 0.6100 & 0.7503 \\
 &  & $H(s_d \rightarrow s_t)$ & 0.8985 & \textbf{0.9613} & \textbf{0.9762} & 0.6160 & 0.7554 \\ \hline
\end{tabular}%
}
\end{table}

Table~\ref{tab:hierarchy} presents the performance comparisons between BERT and the variants of BERT-Joint covering the four different model structures. We have the following observations:
\begin{enumerate}[(1)]\setlength{\itemsep}{0pt}
    \item All variants of BERT-Joint outperform the BERT model, which is not surprising since BERT-Joint takes advantage of additional domain information for intent classification;
    % \item For the two flat structures of $F(\bar{h}; \bar{h})$ and $F(s_d;s_t)$, the latter consistently outperforms the former on all the datasets. This may indicate that the subspace vectors of $s_d$ and $s_t$ can capture effective features from the BERT representation $\bar{h}$ for the corresponding domain and intent classification;
     \item The flat structure of $F(s_d;s_t)$ consistently outperforms $F(\bar{h}; \bar{h})$ on all the datasets, which may indicate that $s_d$ and $s_t$ can capture effective features from the BERT representation $\bar{h}$ for the corresponding domain and intent classification;
    \item The hierarchical structures generally outperform flat structures in terms of accuracy (all) on all the datasets except OOS+ where the structure of $F(s_d; s_t)$ obtains the best accuracy (all), as well as out-of-scope recall and $F_1$;
    \item BERT-Joint is particularly effective in dealing with out-of-scope intent classification. For example, $H(s_d \rightarrow s_t)$ outperforms BERT in terms of $F_1$ by an absolute increase of more than \textbf{10\%} on both \textit{Full} and \textit{Small}. This may be attributed to the domain classification task which also needs to learn how to classify the out-of-domain examples. Such capability is inherited by the intent classifier through feeding the domain representation to the subsequent intent layers and thus the out-of-scope intent classification performance is improved further. 
    % Such out-of-domain signals are propagated back to the intent representation layer and thus improve the out-of-scope intent classification performance by a large margin. 
\end{enumerate}

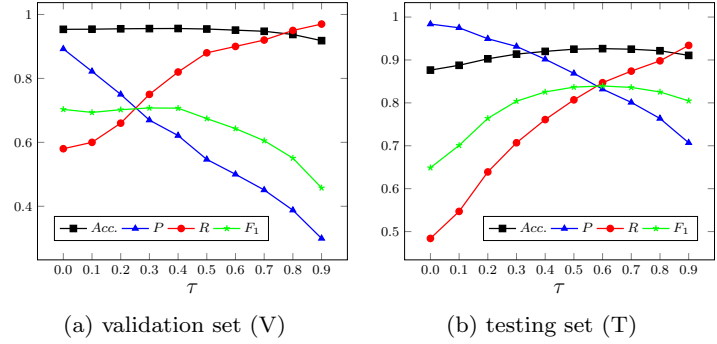
\begin{figure}[htb]
\centering
\begin{subfigure}[b]{0.39\textwidth}
\centering
\resizebox{\linewidth}{!}{ 
\begin{tikzpicture}
  \begin{axis}[
  xlabel=$\tau$, xlabel style={font=\Large},
  legend style={at={(0.4,0.17)},anchor=north}, legend columns=4,
  xticklabels from table={data.dat}{Threshold},xtick=data]
    \addplot[black,thick,mark=square*] table [y=vAcc,x=Index]{data.dat};
    \addlegendentry{$Acc.$}
    \addplot[blue,thick,mark=triangle*] table [y=vP,x=Index]{data.dat};
    \addlegendentry{$P$}
    \addplot[red,thick,mark=*] table [y=vR,x=Index]{data.dat};
    \addlegendentry{$R$}
    \addplot[green,thick,mark=star]  table [y=vF1,x=Index]{data.dat};
    \addlegendentry{$F_1$}
  \end{axis}
\end{tikzpicture}}
\caption{validation set (V)}
\end{subfigure}
~
\begin{subfigure}[b]{0.39\textwidth}
\resizebox{\linewidth}{!}{ 
\begin{tikzpicture}
  \begin{axis}[
  xlabel=$\tau$, xlabel style={font=\Large},
  legend style={at={(0.6,0.17)},anchor=north}, legend columns=4,
  xticklabels from table={data.dat}{Threshold},xtick=data]
    \addplot[black,thick,mark=square*] table [y=tAcc,x=Index]{data.dat};
    \addlegendentry{$Acc.$}
    \addplot[blue,thick,mark=triangle*] table [y=tP,x=Index]{data.dat};
    \addlegendentry{$P$}
    \addplot[red,thick,mark=*] table [y=tR,x=Index]{data.dat};
    \addlegendentry{$R$}
    \addplot[green,thick,mark=star]  table [y=tF1,x=Index]{data.dat};
    \addlegendentry{$F_1$}
    \end{axis}
\end{tikzpicture}}
\caption{testing set (T)}
\end{subfigure}
% \vspace{-0.5em}
% \caption{Overall accuracy ($Acc.$) and out-of-scope intent classification performance ($P$, $R$, and $F_1$) on V and T using different threshold values.}
\caption{Overall accuracy ($Acc.$) and out-of-scope intent classification performance ($P$, $R$, and $F_1$) on V and T.}
\label{fig:threshold}
\end{figure}

%%%%%%%%%%%%%%%%%%%%%%%%%%%%%%%%%%%%%%%%%%%
\noindent{\textbf{Threshold-based Post-processing.}}
Figure~\ref{fig:threshold} presents the performance comparisons on the validation dataset (V) and and the testing dataset (T) of OOS with $\tau \in \{0.1, \ldots, 0.9\}$.
It is clear to see that the threshold value $\tau$ affects all the metrics on both V and T, and thus the threshold-based post-processing method provides an effective way to \textit{balance} precision and recall for out-of-scope intent classification.
% With the increase of $\tau$, the accuracy (\textit{Acc.}) improves first and then drops when $\tau>0.4$, while the in-scope accuracy (in) generally decreases along with increasing $\tau$ since the low-probability ($<\tau$) in-scope examples are now misclassified as out-of-scope. 

In Figure~\ref{fig:threshold} (a), with the increase of $\tau$, the accuracy (\textit{Acc.}) improves first and then drops when $\tau > 0.4$, since the low-probability ($<\tau$) in-scope examples are now misclassified as out-of-scope.
As illustrated in Figure~\ref{fig:threshold} (a) and (b) for out-of-scope intent classification, $R$ keeps increasing at the expense of decreasing in $P$, whereas the highest $F_1$ is obtained at $\tau=0.3$ on V and at $\tau=0.6$ on T. Note that T has a much larger number of out-of-scope examples than V and thus requires a larger $\tau$ for better recall.

\begin{figure}[htb]
  \centering
  \includegraphics[width=0.777\linewidth]{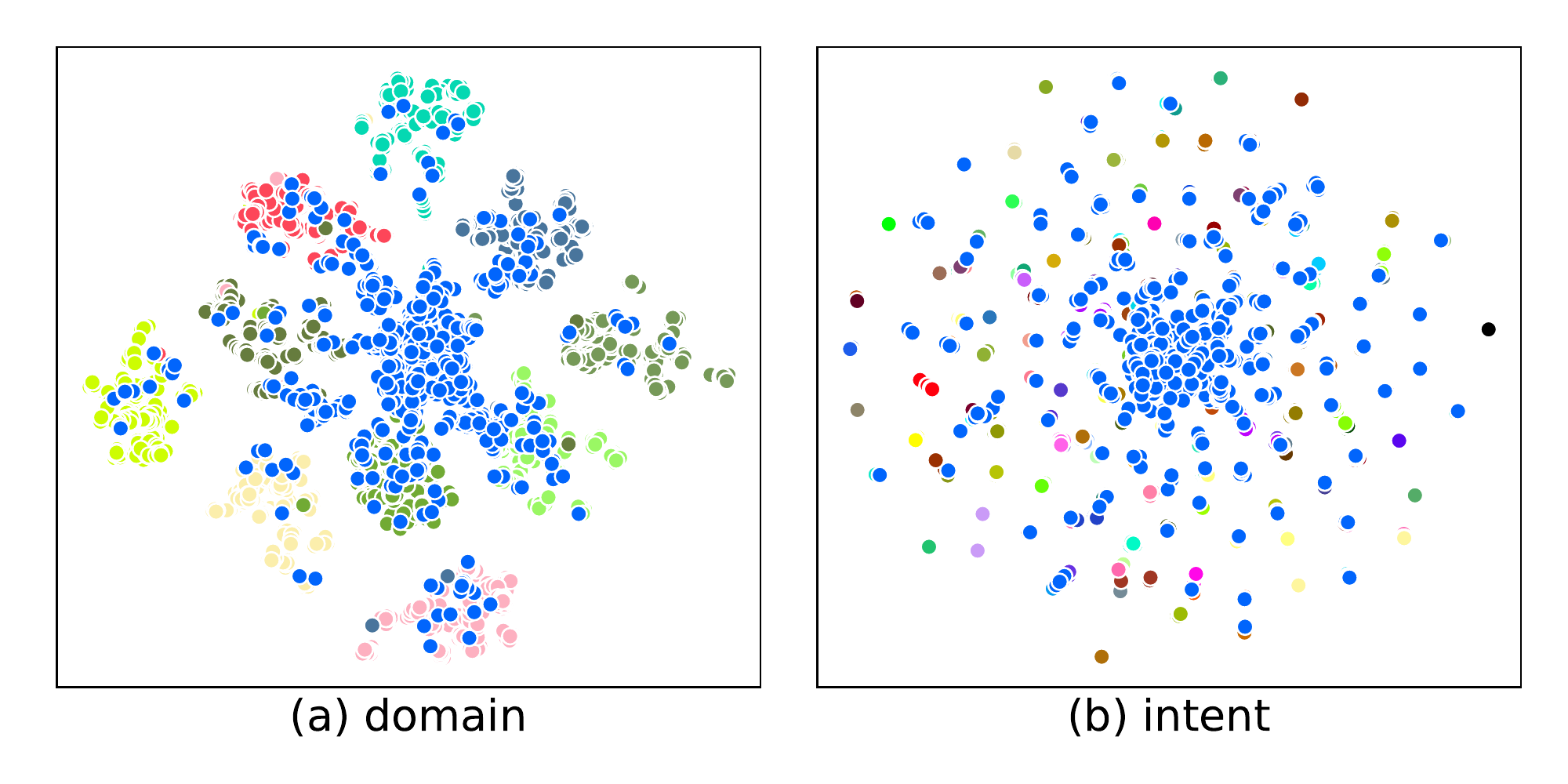}
%   \vspace{-1em}
  \caption{Visualization of domain and intent representations using t-SNE, where each color indicates a domain (intent) and the out-of-scope examples are colored in blue, best viewed when enlarged.}
%   \caption{Visualization of domain and intent representations using t-SNE, where each color indicates a domain (intent) and the out-of-scope examples are colored in blue. Some examples are overlapped, best viewed when enlarged. }
  \label{fig:visualization}
\end{figure}

%%%%%%%%%%%%%%%%%%%%%%%%%%%%%%%%%%%%%%%%%%%%%%%%%
\noindent{\textbf{Representation Visualization.}}
% To analyze the performance of BERT-Joint qualitatively, 
To deepen understanding of the out-of-scope classification problem, we further visualized the domain and intent representations from the test set of OOS using t-SNE \cite{maaten2008visualizing}, which visualizes high-dimensional vectors in a two or three-dimensional map. 
As illustrated in Figure~\ref{fig:visualization}, each color represents a domain (a) or an intent (b). 
The 10 in-scope domains are well separated in Figure~\ref{fig:visualization} (a), so does the 150 in-scope intents in Figure~\ref{fig:visualization} (b). Note that some points are overlapped in Figure~\ref{fig:visualization} due to too many examples and domains/intents, best viewed when enlarged.
% As illustrated in Figure~\ref{fig:visualization}, 
The out-of-scope examples are mainly located in the same \textit{blue} cluster in both (a) and (b), but quite a few out-of-scope examples are distributed across different domains or intents.
% The 10 in-scope domains are well separated in Figure~\ref{fig:visualization} (a), so does the 150 in-scope intents in Figure~\ref{fig:visualization} (b). Note that some points are overlapped in Figure~\ref{fig:visualization} due to too many examples and domains/intents, best viewed when enlarged.
This explains why it is difficult to classify the out-of-scope examples, and why the simple threshold-based method gives better performance on out-of-scope intent classification.

% \vspace{-1em}
%%%%%%%%%%%%%%%%%%%

%%%%%%%%%%%%%%%%%%
%%%%%%%%%%%%%%%%%%%%
% \vspace{-1em}
\section{Conclusion}
This paper presents a novel hierarchical joint model based on BERT for out-of-scope domain and intent classification.
The proposed model allows sharing of supervised signals between both classification tasks and introduces a structural bias to enable hierarchical representation learning from the pre-trained BERT representations.
We empirically show that the model outperforms existing methods in terms of accuracy as well as out-of-scope recall and $F_1$ by a large margin on all the variants of the OOS dataset.
These observations serve to illustrate the effectiveness of joint modeling and hierarchical structure of the model particularly in out-of-scope intent classification.
Furthermore, we show that a threshold-based post-processing method improves the performance further and allows to effectively balance precision and recall in out-of-scope intent classification.
% For future work, it will be worthwhile to investigate the extension to a slot filling task into the hierarchical multi-task learning framework for natural language understanding in task-oriented dialog systems.
%%%%%%%%%%%%%%%%%%

\bibliographystyle{IEEEtran}
\bibliography{references}

\begin{thebibliography}{10}
\providecommand{\url}[1]{#1}
\csname url@samestyle\endcsname
\providecommand{\newblock}{\relax}
\providecommand{\bibinfo}[2]{#2}
\providecommand{\BIBentrySTDinterwordspacing}{\spaceskip=0pt\relax}
\providecommand{\BIBentryALTinterwordstretchfactor}{4}
\providecommand{\BIBentryALTinterwordspacing}{\spaceskip=\fontdimen2\font plus
\BIBentryALTinterwordstretchfactor\fontdimen3\font minus
  \fontdimen4\font\relax}
\providecommand{\BIBforeignlanguage}[2]{{%
\expandafter\ifx\csname l@#1\endcsname\relax
\typeout{** WARNING: IEEEtran.bst: No hyphenation pattern has been}%
\typeout{** loaded for the language `#1'. Using the pattern for}%
\typeout{** the default language instead.}%
\else
\language=\csname l@#1\endcsname
\fi
#2}}
\providecommand{\BIBdecl}{\relax}
\BIBdecl

\bibitem{liu2019review}
J.~Liu, Y.~Li, and M.~Lin, ``Review of intent detection methods in the
  human-machine dialogue system,'' in \emph{Journal of Physics: Conference
  Series}, vol. 1267, no.~1.\hskip 1em plus 0.5em minus 0.4em\relax IOP
  Publishing, 2019, p. 012059.

\bibitem{larson2019evaluation}
\BIBentryALTinterwordspacing
S.~Larson, A.~Mahendran, J.~J. Peper, C.~Clarke, A.~Lee, P.~Hill, J.~K.
  Kummerfeld, K.~Leach, M.~A. Laurenzano, L.~Tang, and J.~Mars, ``An evaluation
  dataset for intent classification and out-of-scope prediction,'' in
  \emph{Proceedings of the 2019 Conference on Empirical Methods in Natural
  Language Processing and the 9th International Joint Conference on Natural
  Language Processing (EMNLP-IJCNLP)}, 2019. [Online]. Available:
  \url{https://www.aclweb.org/anthology/D19-1131}
\BIBentrySTDinterwordspacing

\bibitem{hendrycks2017baseline}
D.~Hendrycks and K.~Gimpel, ``A baseline for detecting misclassified and
  out-of-distribution examples in neural networks,'' in \emph{Proceedings of
  International Conference on Learning Representations}, 2017.

\bibitem{lin2019deep}
\BIBentryALTinterwordspacing
T.-E. Lin and H.~Xu, ``Deep unknown intent detection with margin loss,'' in
  \emph{Proceedings of the 57th Annual Meeting of the Association for
  Computational Linguistics}.\hskip 1em plus 0.5em minus 0.4em\relax Florence,
  Italy: Association for Computational Linguistics, Jul. 2019, pp. 5491--5496.
  [Online]. Available: \url{https://www.aclweb.org/anthology/P19-1548}
\BIBentrySTDinterwordspacing

\bibitem{wu2020tod}
C.-S. Wu, S.~Hoi, R.~Socher, and C.~Xiong, ``Tod-bert: Pre-trained natural
  language understanding for task-oriented dialogues,'' \emph{arXiv preprint
  arXiv:2004.06871}, 2020.

\bibitem{meng1999believe}
H.~M. Meng, W.~Lam, and C.~Wai, ``To believe is to understand,'' in
  \emph{Proceedings of the 6th European Conference on Speech Communication and
  Technology}, 1999.

\bibitem{lin2019post}
T.-E. Lin and H.~Xu, ``A post-processing method for detecting unknown intent of
  dialogue system via pre-trained deep neural network classifier,''
  \emph{Knowledge-Based Systems}, vol. 186, p. 104979, 2019.

\bibitem{xu2013convolutional}
P.~Xu and R.~Sarikaya, ``Convolutional neural network based triangular {CRF}
  for joint intent detection and slot filling,'' in \emph{2013 IEEE workshop on
  automatic speech recognition and understanding}.\hskip 1em plus 0.5em minus
  0.4em\relax IEEE, 2013, pp. 78--83.

\bibitem{guo2014joint}
D.~Guo, G.~Tur, W.-t. Yih, and G.~Zweig, ``Joint semantic utterance
  classification and slot filling with recursive neural networks,'' in
  \emph{2014 IEEE Spoken Language Technology Workshop (SLT)}.\hskip 1em plus
  0.5em minus 0.4em\relax IEEE, 2014, pp. 554--559.

\bibitem{liu2016attention}
\BIBentryALTinterwordspacing
B.~Liu and I.~Lane, ``Attention-based recurrent neural network models for joint
  intent detection and slot filling,'' in \emph{Interspeech 2016}, 2016, pp.
  685--689. [Online]. Available:
  \url{http://dx.doi.org/10.21437/Interspeech.2016-1352}
\BIBentrySTDinterwordspacing

\bibitem{zhang2016joint}
X.~Zhang and H.~Wang, ``A joint model of intent determination and slot filling
  for spoken language understanding,'' in \emph{Proceedings of the Twenty-Fifth
  International Joint Conference on Artificial Intelligence}, 2016, pp.
  2993--2999.

\bibitem{kim2017onenet}
Y.-B. Kim, S.~Lee, and K.~Stratos, ``Onenet: Joint domain, intent, slot
  prediction for spoken language understanding,'' in \emph{2017 IEEE Automatic
  Speech Recognition and Understanding Workshop (ASRU)}.\hskip 1em plus 0.5em
  minus 0.4em\relax IEEE, 2017, pp. 547--553.

\bibitem{goo2018slot}
C.-W. Goo, G.~Gao, Y.-K. Hsu, C.-L. Huo, T.-C. Chen, K.-W. Hsu, and Y.-N. Chen,
  ``Slot-gated modeling for joint slot filling and intent prediction,'' in
  \emph{Proceedings of the 2018 Conference of the North American Chapter of the
  Association for Computational Linguistics: Human Language Technologies,
  Volume 2 (Short Papers)}, 2018, pp. 753--757.

\bibitem{wang2018bi}
Y.~Wang, Y.~Shen, and H.~Jin, ``A bi-model based rnn semantic frame parsing
  model for intent detection and slot filling,'' in \emph{NAACL-HLT (2)}, 2018.

\bibitem{chen2019bert}
Q.~Chen, Z.~Zhuo, and W.~Wang, ``Bert for joint intent classification and slot
  filling,'' \emph{arXiv preprint arXiv:1902.10909}, 2019.

\bibitem{zhang2019joint}
\BIBentryALTinterwordspacing
C.~Zhang, Y.~Li, N.~Du, W.~Fan, and P.~Yu, ``Joint slot filling and intent
  detection via capsule neural networks,'' in \emph{Proceedings of the 57th
  Annual Meeting of the Association for Computational Linguistics}.\hskip 1em
  plus 0.5em minus 0.4em\relax Florence, Italy: Association for Computational
  Linguistics, Jul. 2019, pp. 5259--5267. [Online]. Available:
  \url{https://www.aclweb.org/anthology/P19-1519}
\BIBentrySTDinterwordspacing

\bibitem{hakkani2016multi}
D.~Hakkani-T{\"u}r, G.~Tur, A.~Celikyilmaz, Y.-N. Chen, J.~Gao, L.~Deng, and
  Y.-Y. Wang, ``Multi-domain joint semantic frame parsing using bi-directional
  rnn-lstm,'' \emph{Interspeech 2016}, pp. 715--719, 2016.

\bibitem{kim2018joint}
J.-K. Kim and Y.-B. Kim, ``Joint learning of domain classification and
  out-of-domain detection with dynamic class weighting for satisficing false
  acceptance rates,'' \emph{arXiv preprint arXiv:1807.00072}, 2018.

\bibitem{sanh2019hierarchical}
V.~Sanh, T.~Wolf, and S.~Ruder, ``A hierarchical multi-task approach for
  learning embeddings from semantic tasks,'' in \emph{Proceedings of the AAAI
  Conference on Artificial Intelligence}, vol.~33, 2019, pp. 6949--6956.

\bibitem{braun2017evaluating}
D.~Braun, A.~H. Mendez, F.~Matthes, and M.~Langen, ``Evaluating natural
  language understanding services for conversational question answering
  systems,'' in \emph{Proceedings of the 18th Annual SIGdial Meeting on
  Discourse and Dialogue}, 2017, pp. 174--185.

\bibitem{coucke2018snips}
A.~Coucke, A.~Saade, A.~Ball, T.~Bluche, A.~Caulier, D.~Leroy, C.~Doumouro,
  T.~Gisselbrecht, F.~Caltagirone, T.~Lavril \emph{et~al.}, ``Snips voice
  platform: an embedded spoken language understanding system for
  private-by-design voice interfaces,'' \emph{arXiv preprint arXiv:1805.10190},
  2018.

\bibitem{liu2019benchmarking}
X.~Liu, A.~Eshghi, P.~Swietojanski, and V.~Rieser, ``Benchmarking natural
  language understanding services for building conversational agents,'' in
  \emph{Proceedings of the Tenth International Workshop on Spoken Dialogue
  Systems Technology (IWSDS)}, 2019.

\bibitem{yilmaz2020kloos}
E.~H. Yilmaz and C.~Toraman, ``{KLOOS: KL Divergence-based Out-of-Scope Intent
  Detection in Human-to-Machine Conversations},'' in \emph{Proceedings of the
  43rd International ACM SIGIR Conference on Research and Development in
  Information Retrieval}, 2020, pp. 2105--2108.

\bibitem{aliannejadi2019asking}
M.~Aliannejadi, H.~Zamani, F.~Crestani, and W.~B. Croft, ``Asking clarifying
  questions in open-domain information-seeking conversations,'' in
  \emph{Proceedings of the 42nd international acm sigir conference on research
  and development in information retrieval}, 2019, pp. 475--484.

\bibitem{hashimoto2017joint}
\BIBentryALTinterwordspacing
K.~Hashimoto, C.~Xiong, Y.~Tsuruoka, and R.~Socher, ``A joint many-task model:
  Growing a neural network for multiple {NLP} tasks,'' in \emph{Proceedings of
  the 2017 Conference on Empirical Methods in Natural Language
  Processing}.\hskip 1em plus 0.5em minus 0.4em\relax Copenhagen, Denmark:
  Association for Computational Linguistics, Sep. 2017, pp. 1923--1933.
  [Online]. Available: \url{https://www.aclweb.org/anthology/D17-1206}
\BIBentrySTDinterwordspacing

\bibitem{devlin2019bert}
J.~Devlin, M.-W. Chang, K.~Lee, and K.~Toutanova, ``{BERT: Pre-training of Deep
  Bidirectional Transformers for Language Understanding},'' in
  \emph{Proceedings of the 2019 Conference of the North American Chapter of the
  Association for Computational Linguistics: Human Language Technologies,
  Volume 1 (Long and Short Papers)}, 2019, pp. 4171--4186.

\bibitem{he2016deep}
K.~He, X.~Zhang, S.~Ren, and J.~Sun, ``Deep residual learning for image
  recognition,'' in \emph{Proceedings of the IEEE conference on computer vision
  and pattern recognition}, 2016, pp. 770--778.

\bibitem{ba2016layer}
J.~L. Ba, J.~R. Kiros, and G.~E. Hinton, ``Layer normalization,'' \emph{arXiv
  preprint arXiv:1607.06450}, 2016.

\bibitem{vaswani2017attention}
A.~Vaswani, N.~Shazeer, N.~Parmar, J.~Uszkoreit, L.~Jones, A.~N. Gomez,
  {\L}.~Kaiser, and I.~Polosukhin, ``Attention is all you need,'' in
  \emph{Advances in neural information processing systems}, 2017, pp.
  5998--6008.

\bibitem{loshchilov2017decoupled}
I.~Loshchilov and F.~Hutter, ``Decoupled weight decay regularization,''
  \emph{arXiv preprint arXiv:1711.05101}, 2017.

\bibitem{paszke2019pytorch}
A.~Paszke, S.~Gross, F.~Massa, A.~Lerer, J.~Bradbury, G.~Chanan, T.~Killeen,
  Z.~Lin, N.~Gimelshein, L.~Antiga \emph{et~al.}, ``{PyTorch}: An imperative
  style, high-performance deep learning library,'' in \emph{Advances in neural
  information processing systems}, 2019, pp. 8026--8037.

\bibitem{maaten2008visualizing}
L.~v.~d. Maaten and G.~Hinton, ``Visualizing data using {t-SNE},''
  \emph{Journal of machine learning research}, vol.~9, no. Nov, pp. 2579--2605,
  2008.

\end{thebibliography}

\end{document}